\documentclass[letterpaper]{article}
\usepackage{aaai}
\usepackage{times}
\usepackage{helvet}
\usepackage{courier}
\usepackage{amsthm}

\title{Don't Fear the Reaper: Refuting Bostrom's Superintelligence Argument}
\author{Sebastian Benthall\\
UC Berkeley\\
School of Information\\}
\begin{document}
\maketitle

\begin{abstract}
\begin{quote}
In recent years prominent intellectuals have raised ethical concerns about
the consequences of artificial intelligence.
One concern is that an autonomous agent might modify itself to become "superintelligent" and, in supremely effective pursuit of poorly specified goals, destroy all of humanity.
This paper considers and rejects the possibility of this outcome.
We argue that this scenario depends on an agent's ability to 
rapidly improve its ability to predict its environment through
self-modification.
Using a Bayesian model of a reasoning agent, we show that there are important limitations to how an agent may improve its predictive ability through self-modification alone.
We conclude that concern about this artificial intelligence outcome is misplaced and better directed at policy questions around data access
and storage.
\end{quote}
\end{abstract}

\noindent The appetite of the public and prominent intellectuals for the study of the ethical implications
of artificial intelligence has increased in recent years.
One captivating possibility is that artificial intelligence research might result in a
`superintelligence' that puts humanity at risk.
\cite{moonshine} has called for AI researchers to consider this possibility seriously because, however unlikely,
its mere possibility is grave.

\cite{super} argues for the importance of considering
the risks of artificial intelligence as a research agenda.
For Bostrom, the potential risks of artificial intelligence are not just at the
scale of industrial mishaps or weapons of mass destruction.
Rather, Bostrom argues that artificial intelligence has the potential to
threaten humanity as a whole and determine the fate of the universe.
We approach this grand thesis with a measure of skepticism.
Nevertheless, we hope that by elucidating the argument and considering
potential objections in good faith, we can get a better grip on the
realistic ethical implications of artificial intelligence.

This paper is in that spirit. We consider the argument for this AI doomsday scenario proposed by Bostrom
\cite{super}.
Section 1 summarizes Bostrom's argument and motivates the work of the rest of the paper.
In focuses on the conditions of an ``intelligence explosion'' that would lead to a dominant machine
intelligence averse to humanity.
Section 2 argues that rather than speculating broadly about general artificial intelligence,
we can predict outcomes of artificial intelligence by considering more narrowly a few tasks that
are essential to instrumental reasoning.
Section 3 considers recalcitrance, the resistance of a system to improvements to its own intelligence,
and the ways it can limit intelligence explosion.
Section 4 contains an analysis of the recalcitrance of prediction, using a Bayesian model of a
predictive agent. We conclude that prediction is not something an agent can easily improve upon
autonomously.
Section 5 discusses the implication of these findings for further investigation into AI risk.

\section{Bostrom's core argument and definitions}

Bostrom makes a number of claims in the course of his argument which I will
outline here as distinct propositions.

\newtheorem{theorem}{Theorem}
\newtheorem{proposition}[theorem]{Proposition}

\begin{proposition}
A system with sufficient intelligence relative to other intelligent systems will have
a `decisive strategic advantage' and will determine the fate of the world and universe.
\end{proposition}

Concretely, Bostrom accepts human beings, governments, emulated brains, and computers as
potential intelligent systems.
In his implicit model of the world, these agents are in contest with each other.
By Bostrom's definition, a `decisive strategic advantage' is the amount of technological
advantage sufficient ``to achieve complete world domination".
It is beyond the scope of this paper to explore the nuances of Proposition 1.
We will provisionally accept it and focus on the probability that a sufficiently
intelligent system will arise.

\begin{proposition}
An intelligent system is likely to attain a decisive strategic advantage if it undergoes
an `intelligence explosion', a rapidly accelerating rate of intelligence increase.
\end{proposition}

Bostrom never offers a definition of intelligence that is amenable to quantification.
He does leverage quantitative intuitions in the course of his argument when he proposes
the following model of intelligence change.

\begin{proposition}
The rate of change in intelligence is equal to optimization power divided by recalcitrance.
$$\frac{dI}{dt} = \frac{O}{R}$$
Optimization power refers to the effort of improving the intelligence of the system. Recalcitrance refers to the resistance of the system to being improved.
\end{proposition}

\begin{proposition}
If an intelligent system works to improve its own intelligence,
then optimization power will increase with the system's intelligence, leading to rapidly accelerating intelligence increase if recalcitrance is sufficiently low.
\end{proposition}

We will consider more precise versions Proposition 2, 3, and 4 latter in this paper.

Bostrom maintains that an intelligent system will attempt to recursively improve its own
intelligence under very general conditions.

\begin{proposition}
Any intelligent system will have or develop increasing its own intelligence as an instrumental goal
towards other goals.
\end{proposition}

Proposition 5 is a consequence of Bostrom's \textbf{instrumental convergence thesis},  \cite{will}

\begin{quotation}
Several instrumental values can be identified which are convergent in the sense that
their attainment would increase the chances of the agent's goal being realized for 
a wide range of final goals and a wide range of situations, implying that these 
instrumental values are likely to be pursued by a broad spectrum of situated intelligent agents.
\end{quotation}

This thesis is important for Bostrom's line of argument because the threat of AI
comes from its predictably rapid takeoff as a `superintelligence' combined with
the unpredictability of its goals.

\begin{proposition}
A machine intelligence is unlikely to have goals that are aligned with the values of humans.
\end{proposition}

The developing field of value learning in artificial intelligence (cite) has been motivated in
part by concerns akin to Proposition 6.
In Bostrom's work, the problems of machine value misalignment are illustrated by many dystopian
scenarios which we will not go into here.

\begin{proposition}
Recalcitrance is likely to be lower for machine intelligence than for human intelligence because of
the physical properties of computers.
\end{proposition}

The overall picture is a compelling narrative for many.
A machine intelligence research project achieves the ability to modify itself to make itself
more intelligent.
It does so in service of some goal its programmers originally provided (Proposition 5).
Since recalcitrance for improvements to machine intelligence is low (Proposition 7),
it undergoes and intelligence explosion (Proposition 4), gets a decisive strategic advantage
(Proposition 2) and determines the fate of humanity (Proposition 1).
Since the machine's goals are likely misaligned with humanity's (Proposition 6),
artificial intelligence poses a great risk.

Bostrom provides a wide survey of the possibilities surrounding greater-than-human intelligence.
We have outlined the logic of the argument that we believe provides most of the motivational
force behind the book.
In doing so, we have made it easier to verify the logical validity of the argument.
We will continue to analyze this argument with a focus on the role of instrumental goals and
recalcitrance in predicting artificial intelligence related risk.
We will focus on Propositions 2, 3, 4, and 5, leaving other aspects of the argument to
future work.

\section{Intelligence and instrumental tasks}

The use of the term ``intelligence" in the preceding section has been vague.
This is unfortunate and a consequence of some of the vagueness in
discussion of artificial intelligence ethics and risk in Bostrom and elsewhere.
Some of the discourse around the ethics of artificial intelligence anticipates
qualitatively new risks associated with what has been called ``Strong AI" (cite).
One contribution of this paper is to narrow the discussion by showing that these
risks can be understood in terms of well-understood ``narrow" AI tasks.
We anticipate that this narrower framing of the problems of AI risk
will be more tractable.

Bostrom leads with the provocative but fuzzy definition of superintelligence as
``any intellect that greatly exceeds the cognitive performance of humans in virtually
all domains of interest.”
The logic of the argument shows that the ``domains of interest" necessary and sufficient for
intelligence explosion are limited to those that concern intelligence augmentation itself.

Bostrom writes about these domains in two ways.
In one section he discusses the ``cognitive superpowers", domains that would quicken a superintelligence takeoff. These ``superpowers" include: \textit{Intelligence amplification, Strategizing, Social manipulation, Hacking, Technology research, and Economic productivity}.
In another section he discusses ``convergent instrumental values", values that agents with a broad variety of goals would converge on as important to their pursuit of final goals..
These values include: \textit{Self-preservation, Goal-content integrity, Cognitive enhancement, Technological perfection, Resource acquisition.}

There are striking parallels between the ``superpowers" that would hasten takeoff and instrumental values.
``Intelligence amplification" is a superpower, whereas "cognitive enhancement" is an instrumental value.
``Technology research" is a superpower, "technological perfection" is a value.
The danger of intelligence explosion is the danger that an intelligent system will confuse
its power with its motives, in particular when its power and motive are both its own
intelligence in a narrow instrumental sense.
We have captured this aspect of Bostrom's argument in Proposition 5, above.

We note that the motivation of a system to increase its own instrumental intelligence is
necessary but not sufficient for an intelligence explosion.
In addition to being motivated, an intelligent system must be capable of rapidly increasing
its intelligence.
By Proposition 4, this capability will be a function not only of the system's optimization power,
but also its recalcitrance.

The possibility of an intelligence explosion will be restricted specifically by the recalcitrance of
the kinds of tasks that comprise instrumental intelligence.
Narrowing our focus on specific tasks will make the problem of assessing AI risk more tractable
because performance on more narrowly defined tasks is better specified.
As a result, our judgements about the recalcitrance of improvement on those tasks can
be better grounded in statistical and computer science theory, as opposed to being
speculative.

In pursuit of this more narrow and grounded understanding of AI risk, in the next section
we will explore Bostrom's model of intelligence growth in more depth.

\section{Recalcitrance considered}

Bostrom's model of intelligence change depends on two variables, \emph{optimization power}
and \emph{recalcitrance}.
These are presented as components in a qualitative model.
Optimization power is the effort put into improving the intelligence of the system.
Recalcitrance is the resistance of the system to improvement.
While it's desirable to have units in which intelligence, optimization power,
and recalcitrance could be measured, none have been provided by Bostrom.
Nonetheless this model is a useful one for explicating intuitions about
self-modifying intelligence. 

Bostrom's initial formulation of this model is:

$$\frac{dI}{dt} = \frac{O(I)}{R}$$

Bostrom's claim is that for instrumental reasons an intelligent system is likely to invest
some portion of its intelligence back into improving its intelligence.
He introduces a linear model of self-improvement that we will adapt here. 
By assumption we can model $O(I) = \alpha I + \beta$ for some parameters $\alpha$ and $\beta$, where $\alpha$ and $\beta$ are positive and represent the contribution of optimization power by the system itself and external forces (such as a team of researchers), respectively.
If recalcitrance is constant, e.g $R = k$, then we can compute:

$$\frac{dI}{dt} = \frac{\alpha I + \beta}{k}$$

Under these conditions, $I$ will be exponentially increasing in time $t$.
This is the "intelligence explosion" that gives Bostrom's argument so much momentum.
The explosion only gets worse if recalcitrance is below a constant.
Implicitly, Bostrom appears committed to the following additional proposition:

\begin{proposition}
A system whose intelligent is growing exponentially is undergoing an intelligence explosion that will lead to a decisive strategic advantage.
\end{proposition}

We provisionally accept this proposition.
However, it's important to remember that recalcitrance may also be a function of intelligence.
Bostrom does not mention the possibility of recalcitrance \emph{increasing} in intelligence.
Consider the following model where recalcitrance is, like optimization power, linearly increasing in intelligence.

$$\frac{dI}{dt} = \frac{\alpha_o I + \beta_o}{\alpha_r I + \beta_r}$$

Now there are four parameters instead of three.
Note this model is identical to the one above it when $\alpha_r = 0$.
Assuming all these parameters are positive, as $I$ increases the rate of
intelligence growth approaches $\alpha_o / \alpha_r$ from below.
This is linear, not exponential, growth.
In this circumstance, there would be no intelligence explosion and therefore
much less catastrophic AI risk.

There are many plausible reasons why recalcitrance might increase with intelligence levels.
For example, if intelligence improvements vary considerably in the \emph{search cost}
of discovering them, then a system might first collect the ``low hanging fruit" and then
have to resort to searching for harder and harder to reach discoveries.

This is not a decisive argument against intelligence explosion and the 
possibility of a decisively strategic intelligent system.
It is an argument for why considering recalcitrance seriously is important for assessing
the likelihood of such an outcome.
A firmer grip on the problem of predicting future AI risk can be gained by looking
at the recalcitrance of specific instrumental reasoning tasks.
In the next section, we consider specifically the recalcitrance of the general
task of \emph{prediction}.

\section{Recalcitrance of prediction}

Prediction is a very well-studied problem in artificial intelligence and statistics.
Many more specific intelligence tasks can be analyzed as special cases of prediction.
For example, some of Bostrom's ``cognitive superpowers", such as Hacking and
Social Manipulation, are analysable partly as a matter of prediction in the
domains of computer networks and interpersonal interaction.
One reason why prediction is so well-studied is that it is so important \emph{instrumentally}:
skill at prediction is valuable in pursuit of a wide range of other goals.

Prediction is such a critically important part of intelligence that we propose the
following conjecture as an addendum to Bostrom's intelligence explosion argument:

\begin{proposition}
Part of what it means for an intelligent system to improve its own intelligence in a domain (including the domain of improving its own intelligence) is for it to improve its ability to make predictions in that domain.
\end{proposition}

It follows that if the task of prediction is highly recalcitrant, then there will be no
autonomous intelligence explosion.

A benefit of looking at a particular intelligent task is that it allows us to think more
concretely about what it would mean to become more intelligent.
For prediction, we can consider intelligence to be the ability to make good predictions
about the world based on valid inference from data.

We will represent a predicting agent using the Bayesian formulation of statistical inference:

$$P(H|D) = \frac{P(D|H) P(H)}{P(D)}$$

Here, $P(H|D)$ is the posterior probability of a hypothesis $H$ given observed data $D$. If one is following statistically optimal procedure, one can compute this value by taking the prior probability of the hypothesis $P(H)$, multiplying it by the likelihood of the data given the hypothesis $P(D|H)$, and then normalizing this result by dividing by the probability of the data over all models, $P(D) = \sum_{i}P(D|H_i)P(H_i)$.

Statisticians will justifiably argue whether this is the best formulation of prediction. And depending on the specifics of the task, the target value may well be some function of posterior (such as the hypothesis with maximum likelihood) and the overall distribution may be secondary. These are valid objections that we would like to put to one side in order to get across the intuition of an argument.

To the extent that the Bayesian formulation is an accurate representation of the general problem
of prediction, we can analyze its recalcitrance.
We start by enumerating the ways in which an agent might improve its performance on the
prediction task, which is validly computing $P(H|D)$ in such a way that best approximates the truth.

\begin{center}
	\begin{table}
    \begin{tabular}{| l | l | l | l |}
    \hline
    Factor & Performance Bound & Recalcitrance \\ \hline
    Accuracy & Perfect update & $\infty$ at limit \\
    Speed & Hardware limit & $\infty$ at limit \\ 
    Prior & True prior & $\infty$, unalterable \\
    Data & \emph{No bound} & Unknown \\
    \hline
    \end{tabular}
    \caption{Summary of the four factors of prediction performance of an artificially intelligent system, what bounds them, and their effective recalcitrance to self-improvement.}
    \end{table}
\end{center}
\begin{itemize}
\item \textbf{Computational accuracy}. A system can improve its ability to compute the
mathematical function of the Bayesian update. Many widely used statistical inference
algorithms use numerical approximation rather and so it is possible for a system to
improve its algorithm's faithfulness to the mathematical formula that defines its goal.
\item \textbf{Computational speed}. There are faster and slower ways to compute the 
inference formula. An intelligent system could come up with a way to make itself
compute its answer faster.
This might be independent of the accuracy of its answer.
\item \textbf{Prior}. The success of inference depends crucially on the prior 
probability assigned to hypotheses or models. A prior is better when it assigns 
higher probability to the true process that generates observable data, or models 
that are `close' to that true process. 
\item \textbf{Data}. Assuming accurate Bayesian computation, performance at prediction
will depend on the quality of the data used in the inference. Note that "better data" is not
necessarily  the same as ``more data".
If the data that the system learns from is from a biased 
sample of the phenomenon in question, then a successful Bayesian update could 
make its predictions worse, not better. Better data is data that is informative 
with respect to the true process that generated the data.
\end{itemize}

Now that we have enumerate the ways in which an intelligent system may improve its power 
of prediction, we can ask: how recalcitrant are these factors to self-improvement 
by an intelligent system?

\begin{itemize}
\item \textbf{Recalcitrance of accuracy}. It may be possible for a system to inspect itself
and determine ways to modify its own algorithm to make it more accurate at computing
a Bayesian update. However there is a hard limit to improvements of this kind.
No system can compute a Bayesian update more accurately than computing it
\emph{perfectly accurately}. Therefore, in the limit, recalcitrance of computational
accuracy is infinite.
\item \textbf{Recalcitrance of speed}. It is possible for a system to experiment with novel
algorithms and select those that are provably faster than its current ones,
or which perform better on benchmark tests.
However, once again there is a hard limit to the speed of computation: the maximum speed
of the hardware system on which the system is implemented.
Without the ability to increase hardware resources available, an intelligent system will
reach infinite speed recalcitrance in the limit.
\item \textbf{Recalcitrance of the prior}. An intelligent system could modify the parameters of
its expectations independently of the data that it learns from.
But it is essential to the abstraction of the Bayesian agent that the prior encodes whatever
bias the agent has that is not learned from external data.
So strictly speaking, there is no way for an intelligent agent to modify its own prior intelligently.
Intelligence in prediction is a matter of using data intelligently, not being accidentally
gifted with the correct prior beliefs. So the recalcitrance of improving the prior is infinite.
\item \textbf{Recalcitrance of data}. Better data improves performance on prediction. But data collection is not
something an intelligent system can do purely \emph{autonomously}, since it has to interact with 
the phenomenon of interest to get more data.
We cannot make assumptions about the recalcitrance of data collection without modeling the
environment the agent is in.
\end{itemize}

Contrary to the conditions of Bostrom's intelligence explosion scenario, we have identified ways
in which the recalcitrance of prediction, an important instrumental reasoning task, is prohibitively
high.
Purely algorithmic self-improvement is particularly limited.
If we allow a system to improve its own hardware, that allows the system to improve its speed.
Overall performance depends critically on data collection.
Neither hardware expansion nor data collection is a feature of the intelligent system alone,
but rather the possibility of these depends on the context in which the system operates.
If, for example, there are increasing search costs for the intelligent system as it seeks out
new data and hardware improvements, that would imply an increase in recalcitrance as a function
of intelligence.
As we have seen in the previous section, this sort of dependence of recalcitrance on intelligence
can mean that the probability of an intelligence explosion is negligible.

\section{Discussion and directions for future work}

We have explicated the logic of one argument for concern about risk from artificial intelligence.
This argument concerns the possibility that an autonomous intelligent system modifies itself,
undergoes an intelligence explosion, and takes over the world in a way that is adverse to human
interests.

In our analysis, we discover that at the core of the argument are several claims that are much
more narrow and tractable than appear on the surface.
In particular, we can get a grip on the problem of predicting the behavior of self-modifying
intelligent systems by focusing on instrumental reasoning tasks and their susceptibility to
autonomous self-improvement.
If we can show that recalcitrance on these tasks is predictably high, we can dismiss the
probability of an intelligence explosion as being negligible.

To demonstrate how such an analysis could work, we analyzed the recalcitrance of prediction,
using a Bayesian model of a predictive agent.
We found that the barriers to recursive self-improvement through algorithmic changes is
prohibitively high for an intelligence explosion.
Rather, an intelligent system attempting to improve its own abilities of prediction would
need foremost to acquire faster hardware and better data.

The recalcitrance of acquiring faster hardware and better data depend not just on the
intelligence of the system, but also on the environment.
If an environment imposes variable search and acquisition costs for hardware and data,
we would expect recalcitrance of these improvements to increase with intelligence, which
would curtail an intelligence explosion.

While not a decisive argument against the possibility of an intelligence explosion,
the preceding arguments do suggest that those concerned with the ethical implications of
the future of AI should put their attention elsewhere.
If intelligent systems engage in intelligence-expanding activities but in a
non-explosive way, that suggests that the probably outcomes for the future of AI
will be best modelled as multi-agent systems competing for cognitive resources rather
than as a single, decisively controlling agent.

If intelligence growth is limited by data and hardware, not by advancement in artificial intelligence
algorithms, that also suggests that AI researchers may not be in the best position to mitigate
the risks of artificial intelligence.
Rather, regulators controlling the use of generic computing hardware and data storage may be more important to determining the future of artificial intelligence than those that design algorithms.

%%%%%%%%%%
% References and End of Paper
\bibliographystyle{aaai}
\bibliography{rec.bib}

\end{document}